\documentclass[Afour,sageh,times]{sagej}

\usepackage{moreverb,url}

\usepackage[colorlinks,bookmarksopen,bookmarksnumbered,citecolor=red,urlcolor=red]{hyperref}

\usepackage{graphics} 
\usepackage{graphicx}
\usepackage{amsmath,nccmath} 
\usepackage{amssymb}  
\usepackage{acronym}
\usepackage{mathrsfs}
\usepackage{nicefrac}
\usepackage{subfigure}
\usepackage{pifont}  
\usepackage{afterpage}  

\usepackage{tikz}
\usetikzlibrary{calc, arrows, shapes, positioning}
\usepackage{pgfplots}

\newcommand\BibTeX{{\rmfamily B\kern-.05em \textsc{i\kern-.025em b}\kern-.08em
T\kern-.1667em\lower.7ex\hbox{E}\kern-.125emX}}

\newcommand{\cmark}{\ding{51}}%
\newcommand{\xmark}{\ding{55}}%
\newcommand{\mbf}[1]{\mathbf{#1}}
\newcommand{\mbs}[1]{\boldsymbol{#1}}
\newcommand{\bbm}{\begin{bmatrix}}
\newcommand{\ebm}{\end{bmatrix}}
\gdef\x{\mbs{\xi}}

\def\volumeyear{2016}

\begin{document}

\runninghead{Burnett et al.}

\title{Boreas: A Multi-Season Autonomous Driving Dataset}

\author{Keenan Burnett\affilnum{1}, David J. Yoon\affilnum{1}, Yuchen Wu\affilnum{1}, Andrew Zou Li\affilnum{1}, Haowei Zhang\affilnum{1}, Shichen Lu\affilnum{1}, Jingxing Qian\affilnum{1}, Wei-Kang Tseng\affilnum{1}, Andrew Lambert\affilnum{2}, Keith Y.K. Leung\affilnum{2}, \\ Angela P. Schoellig\affilnum{1}, Timothy D. Barfoot\affilnum{1}}

\affiliation{\affilnum{1}University of Toronto \\ \affilnum{2} Applanix, Trimble}


\corrauth{Keenan Burnett, University of Toronto, 4925 Dufferin Street, Toronto, Ontario, Canada.}

\email{keenan.burnett@robotics.utias.utoronto.ca}

\begin{abstract}
	The Boreas dataset was collected by driving a repeated route over the course of one year, resulting in stark seasonal variations and adverse weather conditions such as rain and falling snow. In total, the Boreas dataset includes over 350km of driving data featuring a 128-channel Velodyne Alpha Prime lidar, a 360$^\circ$ Navtech CIR304-H scanning radar, a 5MP FLIR Blackfly S camera, and centimetre-accurate post-processed ground truth poses. Our dataset will support live leaderboards for odometry, metric localization, and 3D object detection. The dataset and development kit are available at \href{https://www.boreas.utias.utoronto.ca}{boreas.utias.utoronto.ca}.
\end{abstract}


\maketitle

\section{Introduction}

To date, autonomous vehicle research and development has focused on achieving sufficient reliability in ideal conditions such as the sunny climates observed in San Francisco, California or Phoenix, Arizona. Adverse weather conditions such as rain and snow remain outside the operational envelope for many of these systems. Additionally, a majority of self-driving vehicles are currently reliant on highly-accurate maps for both localization and perception. These maps are costly to maintain and may degrade as a result of seasonal changes. In order for self-driving vehicles to be deployed safely, these short-comings must be addressed. 

\begin{figure} [t]
	\centering
	
	\begin{tikzpicture} [arrow/.style={>=latex,red, line width=1.25pt}, block/.style={rectangle, draw,
			minimum width=4em, text centered, rounded corners, minimum height=1.25em, line width=1.25pt, inner sep=2.5pt}]
		
		\node[inner sep=0pt] (boreas)
		{\includegraphics[width=0.95\columnwidth]{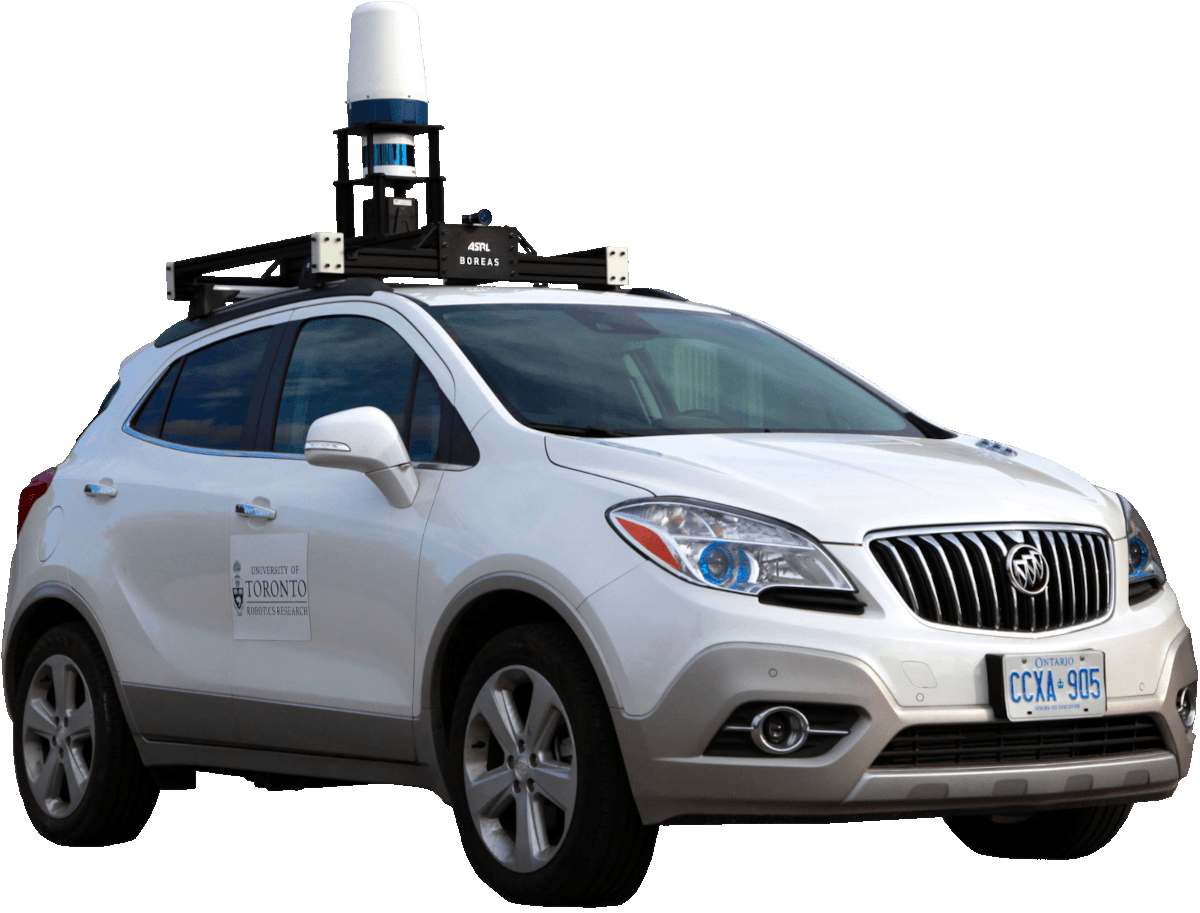}};
		\node (A) [above of=boreas] {};
		\def \L{0.75};
		\coordinate (a1) at ($ (A) + (-1.6, 1.6) $);
		\coordinate (a2) at ($ (a1) + (\L, 0) $);
		\node [block] (radar) at ($ (A) + (-3.5, 1.6) $) {\textbf{360$^{\boldsymbol{\circ}}$ Radar}};
		\coordinate (r1) at ($ (radar.east) + (\L, 0) $);
		\draw[->, arrow] (radar.east) -- (r1) {};
		
		\node [block] (lidar) at ($ (A) + (0.5, 1.4) $) {\textbf{360$^{\boldsymbol{\circ}}$ Lidar}};
		\coordinate (l1) at ($ (lidar.west) + (-\L/2, 0) $);
		\coordinate (l2) at ($ (l1) + (0, -\L/2) $);
		\coordinate (l3) at ($ (l2) + (-\L/2, 0) $);
		\draw[->, arrow] (lidar.west) -- (l1) -- (l2) -- (l3) {};
		
		\node [block] (camera) at ($ (A) + (0.8, 0.6) $) {\textbf{Camera}};
		\coordinate (c1) at ($ (camera.west) + (-\L, 0) $);
		\draw[->, arrow] (camera.west) -- (c1) {};
		
		\node [block] (gps) at ($ (A) + (-3.2, 0.65) $) {\textbf{GNSS/IMU}};
		\coordinate (g1) at ($ (gps.east) + (\L, 0) $);
		\draw[->, arrow] (gps.east) -- (g1) {};
		
	\end{tikzpicture}
	\caption{Our platform, \textit{Boreas}, includes a Velodyne Alpha-Prime (128-beam) lidar, a FLIR Blackfly~S camera, a Navtech CIR304-H radar, and an Applanix POS~LV GNSS-INS.}
	\label{fig:buick}
\end{figure}
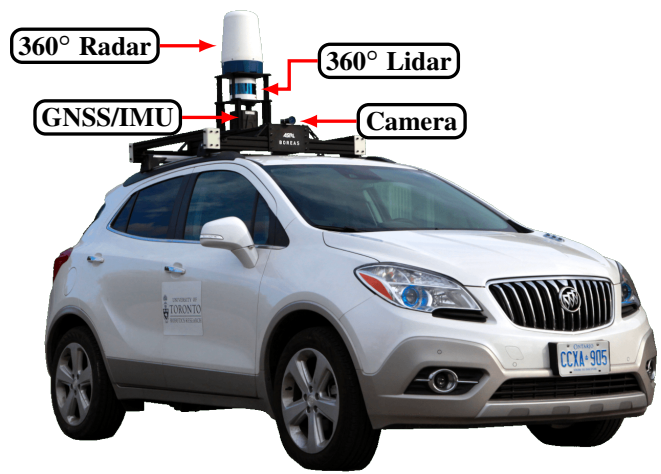

To encourage research in this area, we have created the Boreas dataset, a large multi-modal dataset collected by driving a repeated route over the course of one year. The dataset features over 350km of driving data with stark seasonal variations and multiple sequences with adverse weather such as rain and falling snow. Our data-taking platform, shown in Figure~\ref{fig:buick}, includes a 128-beam lidar, a 5 MP camera, and a $360^\circ$ scanning radar. Globally-consistent centimetre-accurate ground truth poses are obtained by post-processing global navigation satellite system (GNSS), inertial measurement unit (IMU), and wheel encoder data along with a secondary correction subscription. Our dataset will support benchmarks for odometry, metric localization, and 3D object detection.

This dataset may be used to study the effects of seasonal variation on long-term localization. Further, this dataset enables comparisons of vision, lidar, and radar-based mapping and localization pipelines. Comparisons may include the robustness of individual sensing modalities to adverse weather or the resistance to map degradation. 

The main contributions of this dataset are as follows:

\begin{itemize}
	\item Data collected on a repeated route over the course of one year including multiple weather conditions.
	\item A unique, high-quality sensor configuration including a 128-beam lidar and 360$^\circ$ radar.
	\item Post-processed GNSS/IMU data to provide accurate ground truth pose information.
	\item A live and open leaderboard for odometry, metric localization, and 3D object detection.
	\item 3D object labels collected in sunny weather.
\end{itemize}

\renewcommand{\arraystretch}{1.2}
\begin{table*}[ht]
	\small\sf\centering
	\caption{Related datasets. Lead: public leaderboard. Size: For perception datasets, size is given as the number of annotated frames and the number of annotations (3d boxes). GT: ground truth pose source. (A): automotive radar. (N): $360^\circ$ Navtech radar. RTK (Real-Time Kinematic) uses a global positioning system (GPS) base station and differential measurements to improve GPS accuracy. RTX uses data from a global network of tracking stations to calculate corrections. This can be used to achieve cm-level accuracy without a base station \citep{applanixrtx}. $^\dagger$Waymo's Mid-Range, Short-Range proprietary 3D lidar. $^\ddagger$The Oxford Robotcar dataset contains one sequence with snow on the ground but that sequence has no falling snow.}
	
	\resizebox{\textwidth}{!}{\begin{tabular}{ccccccccccc}
			\toprule
			Name & Lead & Size & Camera & Lidar & Radar & GT & Night & Rain & Snow & Seasons\\
			\midrule
			Perception \\ \hline
			
			\begin{tabular}{c} ApolloScape\\ \citet{apolloscape}\end{tabular} & \cmark &  \begin{tabular}{c} 144k\\ 70k boxes\end{tabular}   & 2x9.2MP & 1x64C & \xmark & GPS/IMU & \cmark & \cmark & \xmark & \xmark \\
			
			\begin{tabular}{c} Argoverse\\ \citet{argoverse}\end{tabular} & \cmark &  \begin{tabular}{c} 22k\\ 993k boxes\end{tabular}   & \begin{tabular}{c} 7x2.3MP\\ +2x5MP \end{tabular} & 2x32C & \xmark & GPS/IMU& \cmark & \xmark & \xmark & \xmark \\
			
			\begin{tabular}{c} CADC\\ \citet{cadc}\end{tabular} & \xmark &  \begin{tabular}{c} 7.5k\\ 372k boxes\end{tabular}   & 8x1.3MP & 1x32C & \xmark & GPS/IMU + RTK & \xmark & \xmark & \cmark & \xmark \\
			
			\begin{tabular}{c} KITTI (Object)\\ \citet{kitti}\end{tabular} & \cmark &  \begin{tabular}{c} 15k\\ 200k boxes\end{tabular}   & 4x1.4MP & 1x64C & \xmark & GPS/IMU + RTK & \xmark & \xmark & \xmark & \xmark \\
			
			\begin{tabular}{c} nuScenes\\ \citet{nuscenes}\end{tabular} & \cmark &  \begin{tabular}{c} 40k\\ 1.4M boxes\end{tabular}   & 6x1.4MP & 1x32C & \cmark (A) & \begin{tabular}{c} GPS/IMU\\+ Lidar Loc\end{tabular} & \cmark & \cmark & \xmark & \xmark \\
			
			\begin{tabular}{c} RADIATE\\ \citet{radiate}\end{tabular} & \xmark &  \begin{tabular}{c} 44k\\ 200k boxes\end{tabular}   & 2x0.25MP & 1x32C & \cmark (N) & GPS/IMU & \cmark & \cmark & \cmark & \xmark \\
			
			\begin{tabular}{c} Waymo OD\\ \citet{waymo}\end{tabular} & \cmark &  \begin{tabular}{c} 230k\\ 12M boxes\end{tabular}   & 5x2.5MP & \begin{tabular}{c} 1(MR$^\dagger$)\\ 4(SR$^\dagger$)\end{tabular} & \xmark & GPS/IMU & \cmark & \cmark & \xmark & \xmark \\
			
			\textbf{Boreas-Objects-V1} & \cmark & \begin{tabular}{c} 7.1k \\ 320k boxes \end{tabular} & 1x5MP & 1x128C & \cmark (N) & GPS/IMU & \xmark & \xmark & \xmark & \xmark \\
			
			\hline Localization \\ \hline
			
			\begin{tabular}{c} KITTI (Odometry)\\ \citet{kitti}\end{tabular} & \cmark &  \begin{tabular}{c} 39km\\ 22 seqs\end{tabular}   & 4x1.4MP & 1x64C & \xmark & GPS/IMU + RTK & \xmark & \xmark & \xmark & \xmark \\
			
			\begin{tabular}{c} Complex Urban\\ \citet{complexurban}\end{tabular} & \xmark &  \begin{tabular}{c} 451km\\ 40 seqs\end{tabular}   & 2x1.9MP & \begin{tabular}{c} 2x16C\\+ 2x1C\end{tabular} & \xmark & SLAM & \xmark & \xmark & \xmark & \xmark \\
			
			\begin{tabular}{c} Oxford RobotCar\\ \citet{oxfordrobot}\end{tabular} & \xmark &  \begin{tabular}{c} 1000km\\ 100 seqs\end{tabular}   & \begin{tabular}{c} 3x1.2MP\\ +3x1MP \end{tabular} & \begin{tabular}{c} 1x4C\\+ 2x1C\end{tabular} & \xmark & GPS/IMU + RTK & \cmark & \cmark & \xmark$^\ddagger$ & \cmark \\
			
			\begin{tabular}{c} Oxford Radar\\ \citet{oxfordradar}\end{tabular} & \xmark &  \begin{tabular}{c} 280km\\ 32 seqs\end{tabular}   & \begin{tabular}{c} 3x1.2MP\\ +3x1MP \end{tabular} & \begin{tabular}{c} 2x32C\\+ 2x1C\end{tabular} & \cmark(N) & GPS/IMU + VO & \xmark & \cmark & \xmark & \xmark \\
			
			\begin{tabular}{c} MulRan\\ \citet{mulran}\end{tabular} & \xmark &  \begin{tabular}{c} 124km\\ 12 seqs\end{tabular}   & \xmark & 1x64C & \cmark (N) & SLAM & \xmark & \xmark & \xmark & \xmark \\
			
			\textbf{Boreas} & \cmark &  \begin{tabular}{c} 350km\\ 44 seqs\end{tabular}   & 1x5MP & 1x128C & \cmark (N) & GPS/IMU + RTX & \cmark & \cmark & \cmark & \cmark \\
			
			\bottomrule
	\end{tabular}}
	\label{tab:datasets}
\end{table*}

\section{Related Work}


Many of the published autonomous driving datasets focus on perception, particularly 3D object detection and semantic segmentation of images and lidar pointclouds. However, these datasets tend to lack variation in weather and season. Further, many of these datasets do not provide radar data. Automotive radar sensors are robust to precipitation, dust, and fog thanks to their longer wavelength. For this reason, radar may play a key role in enabling autonomous vehicles to operate in adverse weather. The Boreas dataset addresses these shortcomings by including a $360^\circ$ scanning radar, and data taken during various weather conditions (sun, cloud, rain, night, snow) and seasons. 


Another significant fraction of datasets focus on the problem of localization, usually odometry. The Boreas dataset includes both a high-density lidar (128-beam) and a $360^\circ$ scanning radar. The combination of these sensors and the significant weather variation contained in this dataset enables detailed comparisons between the localization capabilities of these two sensing modalities. This is something that previous datasets were not able to support due to either not having a radar sensor or insufficient weather variation. Furthermore, our post-processed ground truth poses are sufficiently accurate to support a public leaderboard for odometry and metric localization.  Another dataset which focused on adverse weather is RADIATE \citep{radiate}. Whereas RADIATE focused on perception, our dataset focuses on localization. Our dataset is larger and includes repeated traversals of a route with higher-quality localization ground truth. Furthermore, our dataset provides higher resolution radar, lidar, and camera data. For a detailed comparison of related datasets, see Table~\ref{tab:datasets}.

\section{Data Collection}
\begin{figure*}[h]
	\centering
	\includegraphics[width=1.0\textwidth]{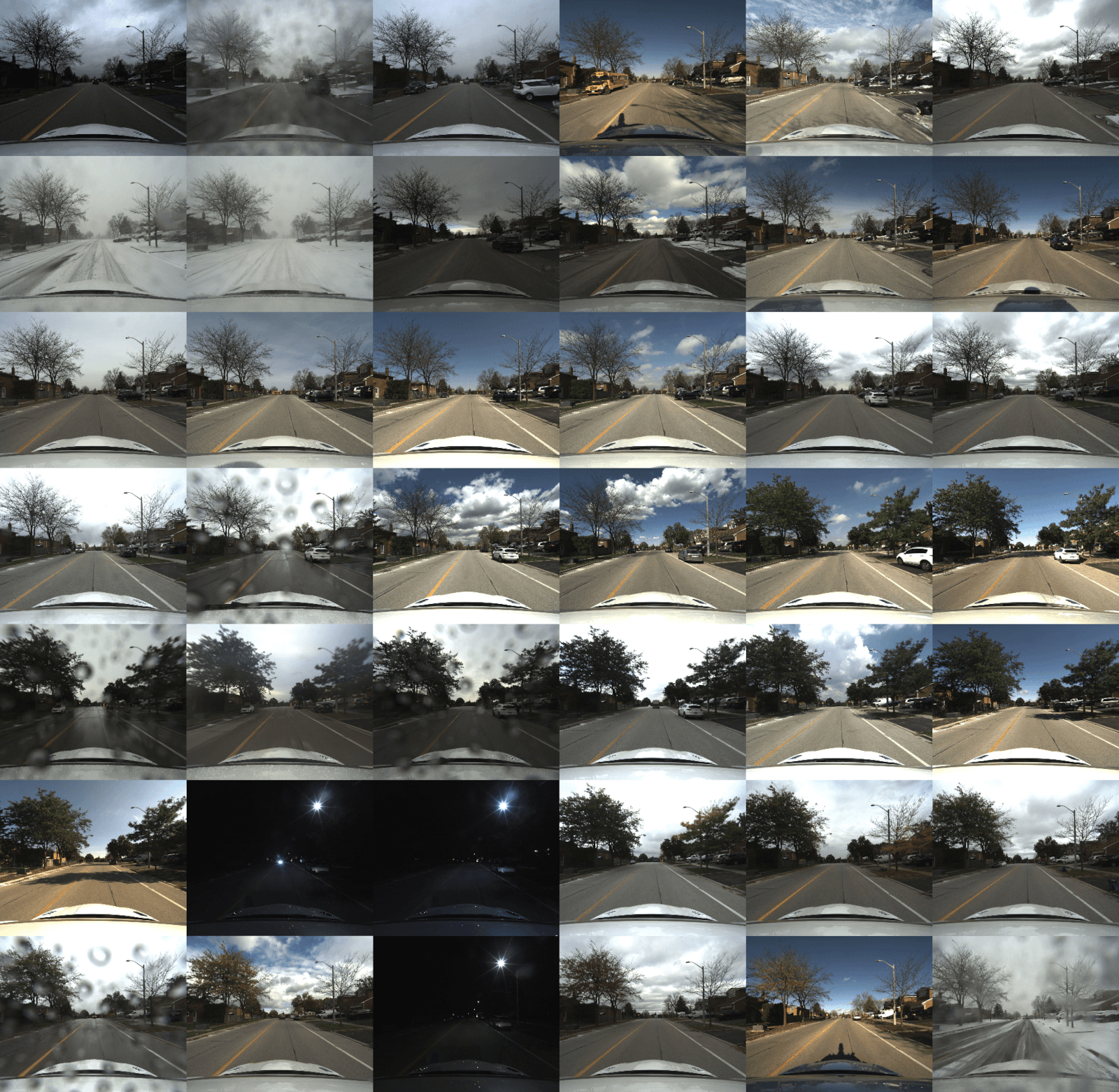}
	\caption{This figure depicts one year of seasonal changes in the Boreas dataset. Each image represents a camera image that was taken on a different day. The sequences are sorted in chronological order from left to right and top to bottom, starting in November, 2020 and finishing in November, 2021. Note that the sequences are not evenly spaced in time.}
	\label{fig:seasons}
\end{figure*}

\begin{figure}[t]
	\centering
	\includegraphics[width=0.9\columnwidth]{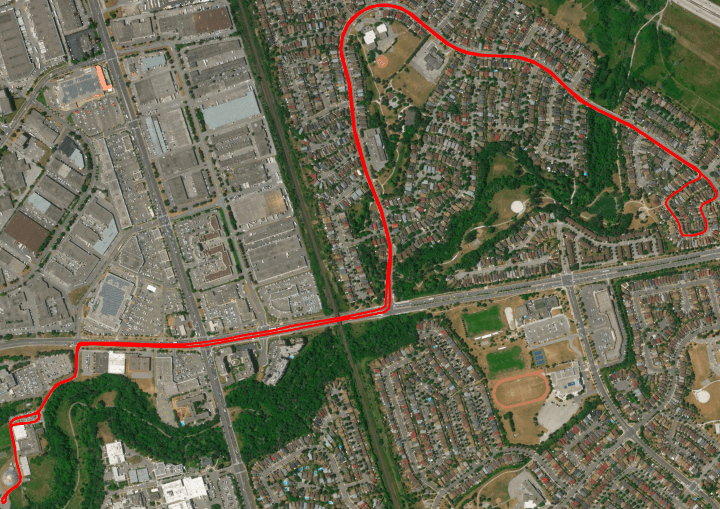}
	\caption{The \href{https://youtu.be/Cay6rSzeo1E}{Glen Shields route} in Toronto, Ontario, Canada. Mapbox satellite data was used to generate this figure.}
	\label{fig:maps}
\end{figure}

\begin{figure}[t]
	\centering
	\includegraphics[width=0.9\columnwidth]{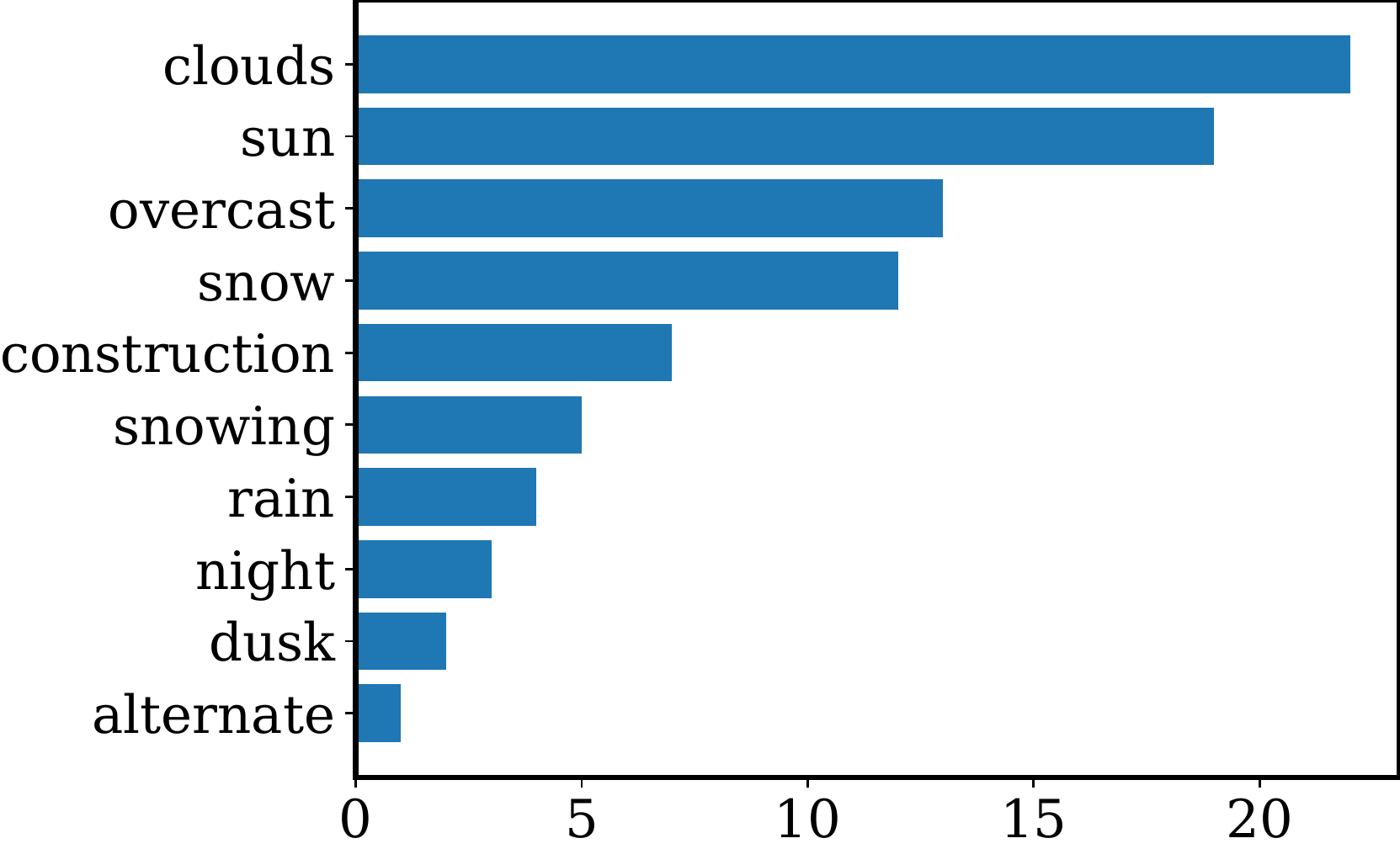}
	\caption{Frequency of metadata tags in the Boreas dataset. Snow: snow is on the ground, snowing: it is actively snowing, alternate: a route other than Glen Shields.}
	\label{fig:meta}
\end{figure}

The majority of the Boreas dataset was collected by driving a repeated route near the University of Toronto over the course of one year. Figure~\ref{fig:seasons} illustrates the seasonal variations that were observed over this time. Figure~\ref{fig:weather} compares camera, lidar, and radar measurements in three distinct weather conditions: falling snow, rain, and sun. The primary repeated route will be referred to as the Glen Shields route and is depicted in Figure~\ref{fig:maps}. Additional routes were also collected as either a single standalone sequence or a small number of repeated traversals. The Glen Shields route can be used for research related to long-term localization while the other routes allow for experiments that test for generalization to previously unseen environments. The frequency of different metadata tags is displayed in Figure~\ref{fig:meta}.



\section{Sensors}

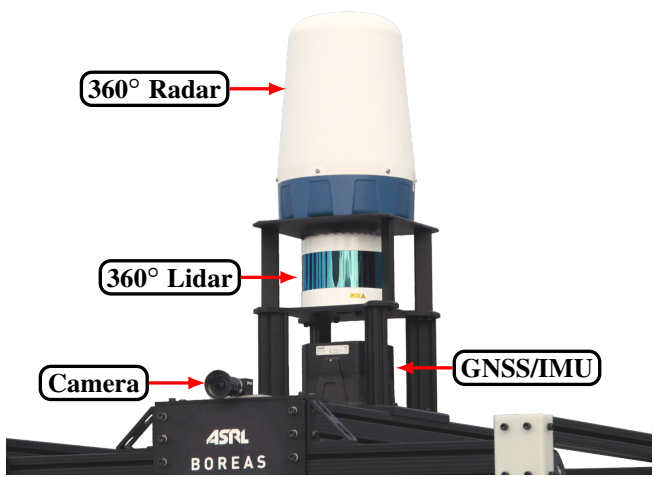
\begin{figure} [t]
	\centering
	
	\begin{tikzpicture} [arrow/.style={>=latex,red, line width=1.25pt}, block/.style={rectangle, draw,
			minimum width=4em, text centered, rounded corners, minimum height=1.25em, line width=1.25pt, inner sep=2.5pt}]
		
		\node[inner sep=0pt] (boreas)
		{\includegraphics[width=\columnwidth]{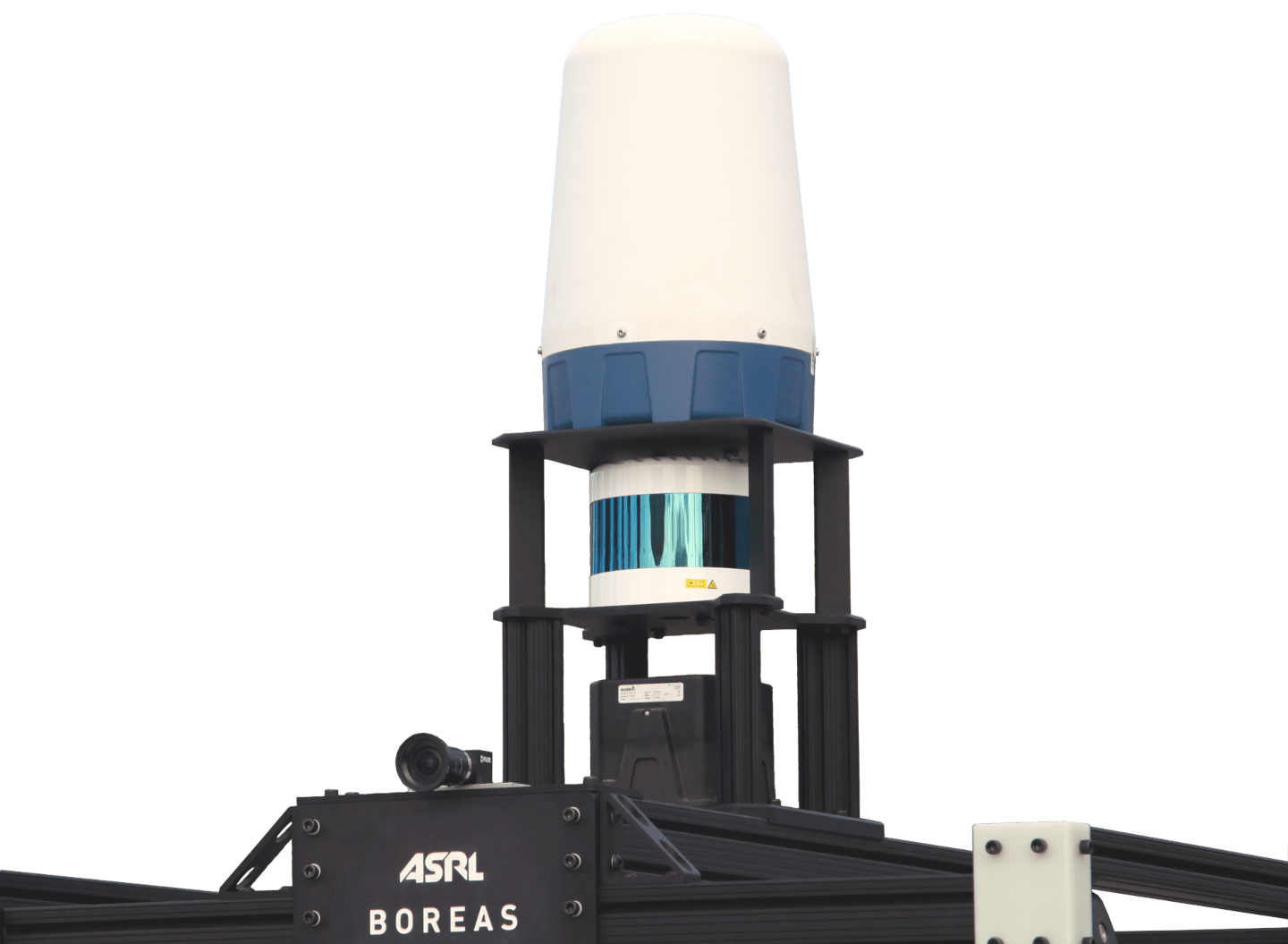}};
		\node (A) [above of=boreas] {};
		\def \L{0.75};
		\node [block] (radar) at ($ (A) + (-2.3, 1.0) $) {\textbf{360$^{\boldsymbol{\circ}}$ Radar}};
		\coordinate (r1) at ($ (radar.east) + (\L, 0) $);
		\draw[->, arrow] (radar.east) -- (r1) {};
		
		\node [block] (lidar) at ($ (A) + (-2.1, -1.5) $) {\textbf{360$^{\boldsymbol{\circ}}$ Lidar}};
		\coordinate (l1) at ($ (lidar.east) + (\L, 0) $);
		\draw[->, arrow] (lidar.east) -- (l1) {};
		
		\node [block] (camera) at ($ (A) + (-3.1, -2.9) $) {\textbf{Camera}};
		\coordinate (c1) at ($ (camera.east) + (\L, 0) $);
		\draw[->, arrow] (camera.east) -- (c1) {};
		
		\node [block] (gps) at ($ (A) + (2.6, -2.7) $) {\textbf{GNSS/IMU}};
		\coordinate (g1) at ($ (gps.west) - (\L, 0) $);
		\draw[->, arrow] (gps.west) -- (g1) {};

		%
		%
		
	\end{tikzpicture}
	\caption{A close-up view of Boreas' sensor configuration.}
	\label{fig:boreas_close}
\end{figure}

\begin{figure}[t]
	\centering
	\includegraphics[width=1.0\columnwidth]{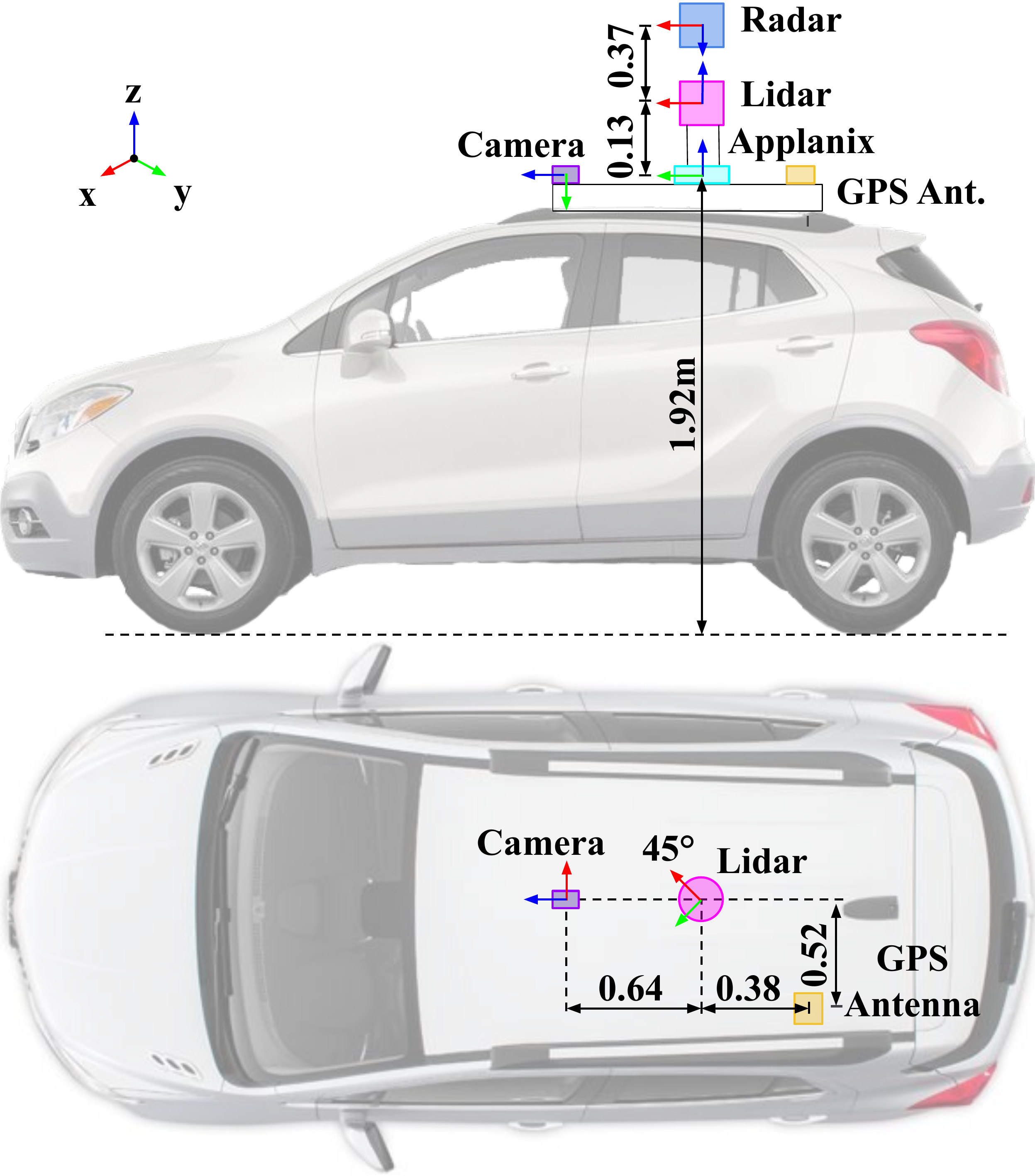}
	\caption{Boreas sensor placement. Distances are given in metres. Measurements shown are approximate. Refer to the calibrated extrinsics contained in the dataset for precise measurements.}
	\label{fig:sensors}
\end{figure}

Table~\ref{tab:sensors} provides detailed specifications for the sensors used in this dataset. Figures~\ref{fig:boreas_close}~and~\ref{fig:sensors} illustrate the placement of the different sensors on Boreas.

\section{Dataset Format}

\subsection{Data Organization}

The Boreas dataset is divided into \textit{sequences}, which include all sensor data and ground truth poses from a single drive. Sequences are identified by the date and time at which they were collected with the format {\ttfamily{boreas-YYYY-MM-DD-HH-MM}}. The data for each sequence is organized as shown in Figure~\ref{fig:data}.

\subsection{Timestamps} \label{sec:time}

The name of each file corresponds to its timestamp. These timestamps are given as UNIX epoch times in microseconds. All sensor timestamps were synchronized to the coordinated universal time (UTC) time reported by the Applanix POS~LV. The Velodyne lidar was synchronized using a standard hardwired connection to the Applanix POS~LV carrying a pulse-per-second (PPS) signal and NMEA messages. The camera was configured to emit a square-wave pulse where the rising edge of each pulse corresponds with the start of a new camera exposure event. The Applanix POS~LV was then configured to receive and timestamp these event signals. Camera timestamps were then corrected in post using the recorded event times and exposure values: $t_{\text{camera}} = t_{\text{event}} + \frac{1}{2}\text{exposure}(t_{\text{event}})$.


\begin{figure*}[h]
	\centering
	\includegraphics[width=0.75\textwidth]{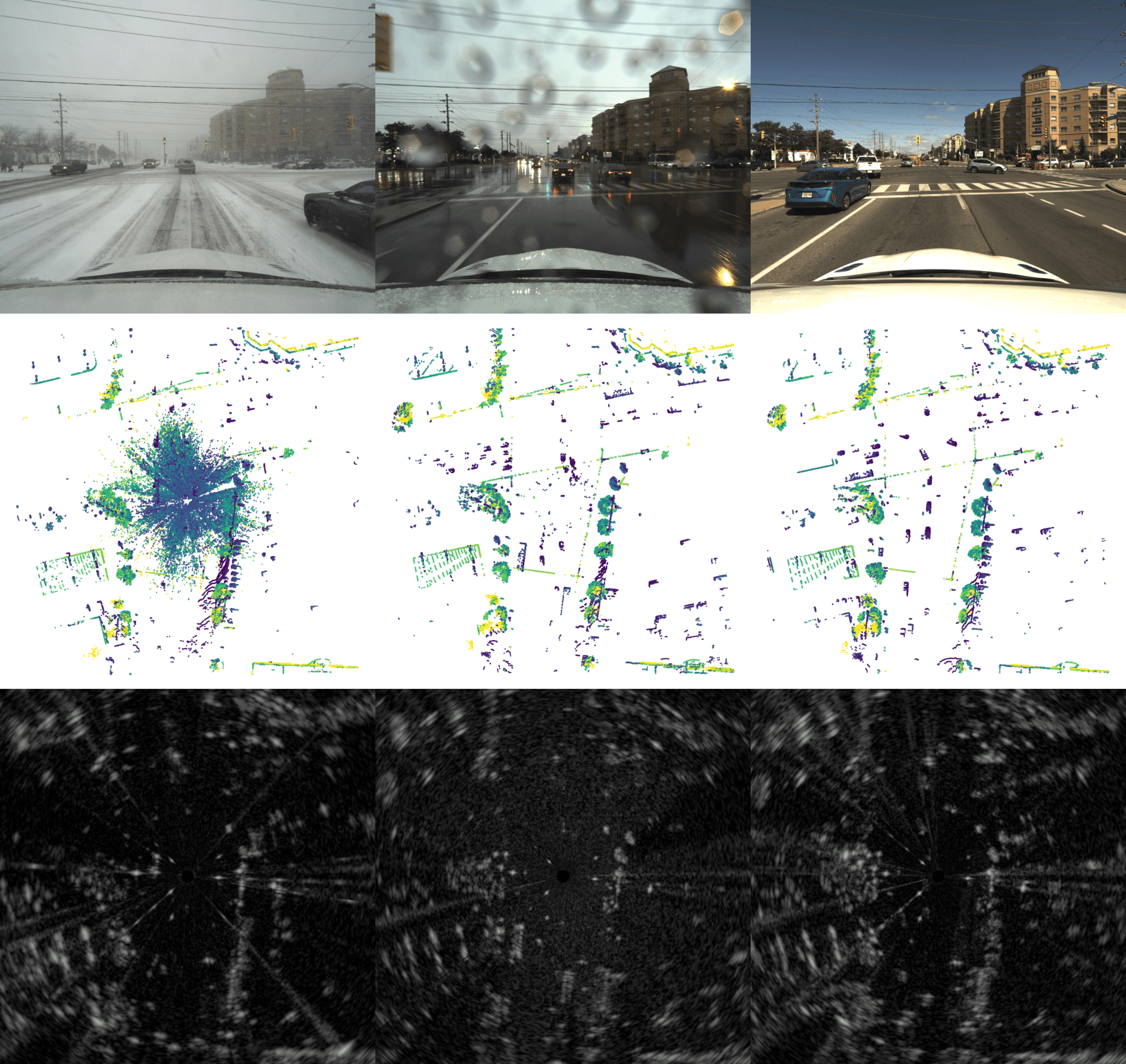}
	\caption{Weather variation in the Boreas dataset. Note that the lidar pointcloud becomes littered with detections associated with snowflakes during falling snow and that the radar data remains relatively unperturbed across the weather conditions.}
	\label{fig:weather}
\end{figure*}

\begin{figure}[b]
	\centering
	\includegraphics[width=1.0\columnwidth]{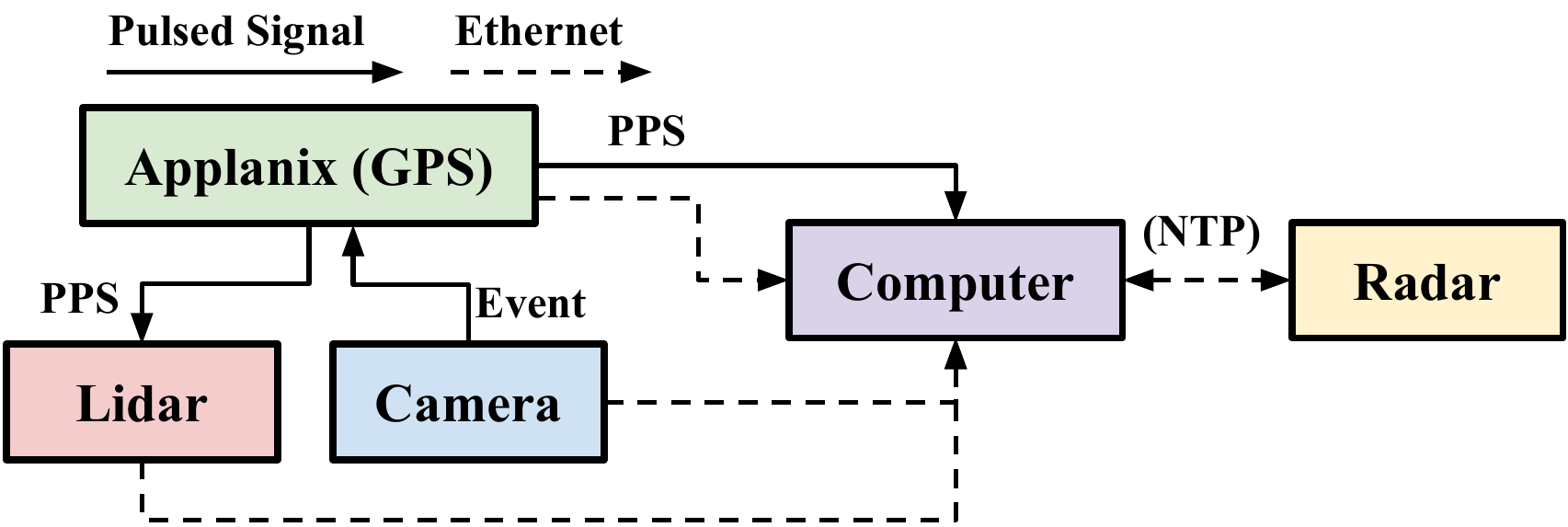}
	\caption{Time synchronization of sensors on Boreas.}
	\label{fig:boreas_time_sync}
\end{figure}



The data-recording computer was synchronized to UTC time in a fashion similar to the Velodyne, using an RS-232 serial cable carrying a PPS signal and NMEA messages. The Navtech radar synchronizes its local clock using network time protocol (NTP). Since the data-recording computer publishing the NTP time is synchronized to UTC time, the radar is thereby also synchronized to UTC time.

For lidar pointclouds, the timestamp corresponds to the temporal middle of the scan. Each lidar point also has a timestamp associated with it. These point times are given in seconds relative to the middle of the scan. For radar scans, the timestamp also corresponds to the the middle of the scan: $\lfloor \frac{M}{2} \rfloor - 1$ where $M$ is the number of azimuths. Each scanned radar azimuth is also timestamped in the same format as the filename, a UNIX epoch time. A diagram of our synchronization setup is shown in Figure~\ref{fig:boreas_time_sync}.

\subsection{File Formats}

Camera images are rectified and anonymized by default. We use Anonymizer to blur license plates and faces \citep{anonymizer}. Images are stored in the commonly-used {\ttfamily{png}} format. Lidar pointclouds are stored in a binary format to minimize storage requirements. Our devkit provides methods for working with these binary formats in both C++ and Python. Each point has six fields: $[x,y,z,i,r,t]$ where $(x,y,z)$ is the position of the point with respect to the lidar, $i$ is the intensity of the reflected infrared signal, $r$ is the ID of the laser that made the measurement, and $t$ the point timestamp explained in Section~\ref{sec:time}. Raw radar scans are stored as 2D polar images: M azimuths x R range bins. We follow Oxford's convention and embed timestamp and encoder information into the first eleven columns (bytes) of each polar radar scan. The first eight columns represent a 64-bit integer, the UNIX epoch timestamp of each azimuth in microseconds. The next two columns represent a 16-bit unsigned integer, the rotational encoder value. The next column is unused but preserved for compatibility with the Oxford format. See \citep{oxfordradar} for further details on the Navtech sensor and this file format. The polar radar scans can be readily converted into a top-down Cartesian representation, as shown in Figure~\ref{fig:weather}, using our devkit.
\newpage

Note that measurements are not synchronous as in other datasets (KITTI, CADC), which means that measurements with the same index do not have the same timestamp. However, given the timestamps and relative pose information, different sensor measurements can still be fused together. Lidar pointclouds are not motion-corrected, but we do provide methods for removing motion distortion in our devkit. Navtech radar scans suffer from both motion distortion and Doppler distortion, \citet{burnett_ral21} and \citet{burnett_rss21} provide methods to compensate for these effects.

\begin{table}[ht]
	\small\sf\centering
	\caption{Sensor specifications. $^\dagger$Position accuracy changes over time as a function of the number of visible satellites. $^\dagger$These numbers represent expected accuracy in nominal conditions. $^\ddagger$Our Navtech radar's firmware was upgraded partway through the project, older sequences have a range resolution of 0.0596m, and a range of 200m.}
	
	\begin{tabular}{ll}
		\toprule
		Sensor & Specifications\\
		\midrule
		
		Applanix & $\bullet$ 2-4cm RTX accuracy (RMS)$^\dagger$ \\
		POS~LV 220 & $\bullet$ 200 Hz \\[1mm]
		
		Navtech CIR304-H & $\bullet$ 0.0438m range solution$^\ddagger$ \\
		Radar & $\bullet$ 0.9$^\circ$ horizontal resolution \\
		& $\bullet$ 250m range$^\ddagger$ \\
		& $\bullet$ 4 Hz \\[1mm]
		
		FLIR Blackfly S & $\bullet$ 2448x2048 (5 MP) \\
		Camera & $\bullet$ 81$^\circ$ HFOV x 71$^\circ$ VFOV \\
		(BFS-U3-51S5C) & $\bullet$ 10 Hz \\[1mm]
		
		Velodyne & $\bullet$ 128 beams \\
		Alpha-Prime & $\bullet$ 0.1$^\circ$ vertical resolution (variable) \\
		Lidar & $\bullet$ 0.2$^\circ$ horizontal resolution \\
		& $\bullet$ 360$^\circ$ HFOV x 40$^\circ$ VFOV  \\
		& $\bullet$ 300m range (10\% reflectivity) \\
		& $\bullet$ $\sim$ 2.2M points/s \\
		& $\bullet$ 10 Hz \\
		\bottomrule
	\end{tabular}
	\label{tab:sensors}
	\vspace{20mm}
\end{table}

\begin{figure}[t]
	\centering
	\includegraphics[width=0.60\columnwidth]{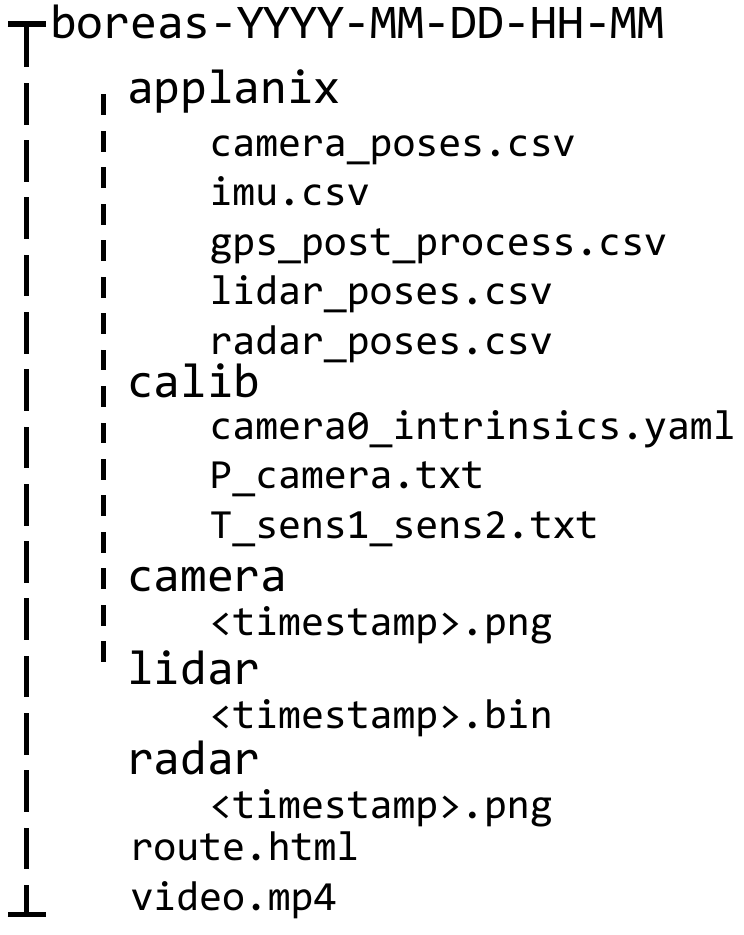}
	\caption{Data organization for a single Boreas sequence.}
	\label{fig:data}
\end{figure}

\section{Ground Truth Poses}



Ground truth poses are obtained by post-processing GNSS, IMU, and wheel encoder measurements along with corrections obtained from an RTX subscription using Applanix's POSPac software suite. Positions and velocities are given with respect to a fixed East-North-Up frame $\text{ENU}_{\text{ref}}$. The position of $\text{ENU}_{\text{ref}}$ is aligned with the first pose of the first sequence ({\ttfamily{boreas-2020-11-26-13-58}}) but the orientation is defined to be tangential to the geoid as defined in the WGS-84 convention such that x points East, y points North, and z points up. For each sequence, {\ttfamily{applanix/gps\_post\_process.csv}} contains the post-processed ground truth in the Applanix frame at 200Hz. We follow the convention used by \citet{barfoot2017state} for describing rotations and $4\times4$ homogeneous transformation matrices. Each sensor frame's ground truth is stored as a row in {\ttfamily{applanix/<sensor>\_poses.csv}} with the following format: $[t, x, y, z, v_x, v_y, v_z, r, p, y, \omega_z, \omega_y, \omega_x]$ where $t$ is the epoch timestamp in microseconds that matches the filename, $\mathbf{r}^{se}_e = [x~y~z]^T$ is the position of the sensor $s$ with respect to $\text{ENU}_{\text{ref}}$ as measured in $\text{ENU}_{\text{ref}}$, $\mathbf{v}^{se}_e = [v_x~v_y~v_z]^T$ is the velocity of the sensor with respect to $\text{ENU}_{\text{ref}}$, $(r, p, y)$ are the roll, pitch, and yaw angles, which can be converted into a rotation matrix between the sensor frame and $\text{ENU}_{\text{ref}}$. $\boldsymbol{\omega}^{se}_s = [\omega_x~\omega_y~\omega_z]^T$ are the angular velocities of the sensor with respect to $\text{ENU}_{\text{ref}}$ as measured in the sensor frame. The pose of the sensor frame is then: $\mbf{T}_{es} = \bbm \mbf{C}_{es} & \mbf{r}_e^{se} \\ \mbf{0}^T & 1 \ebm \in SE(3)$ where $\mbf{C}_{es} = \mbf{C}_1(\text{roll}) \mbf{C}_2(\text{pitch}) \mbf{C}_3(\text{yaw})$ \citep{barfoot2017state}. We also provide post-processed IMU measurements in {\ttfamily{applanix/imu.csv}} at 200Hz in the Applanix frame that include linear acceleration and angular velocity.

The residual root mean square (RMS) position error reported by Applanix is typically less than 5cm in nominal conditions but can be as high as 20-40cm in urban canyons. Figure~\ref{fig:gterr} shows the residual RMS errors resulting from the post-processing conducting by the Applanix POSPac software. The estimated error can change depending on the visibility of satellites. Note that these values represent global estimates and that relative pose estimates are more accurate over short time horizons.


\section{Calibration}

\subsection{Camera Intrinsics}
Camera intrinsics are calibrated using MATLAB's camera calibrator \citep{matlab} and are recorded in {\ttfamily{camera0\_intrinsics.yaml}}. Images located under {\ttfamily{camera/}} have already been rectified. The rectified camera matrix $\mathbf{P}$ is stored in {\ttfamily{P\_camera.txt}}. To project lidar points onto a camera image, we use the pose of the camera $\mathbf{T}_{ec}$ at time $t_c$ and the pose of the lidar $\mathbf{T}_{el}$ at time $t_l$ to compute a transform from the lidar frame to the camera frame given by $\mathbf{T}_{cl} = \mathbf{T}_{ec}^{-1} \mathbf{T}_{el}$. Each point in the lidar frame is then transformed into the camera frame with $\mathbf{x}_c = \mathbf{T}_{cl} \mathbf{x}_l$, where $\mathbf{x}_l = [x~y~z~1]^T$. The projected image coordinates are then obtained using \citep{barfoot2017state}:

\begin{figure}[t]
	\centering
	\includegraphics[width=0.90\columnwidth]{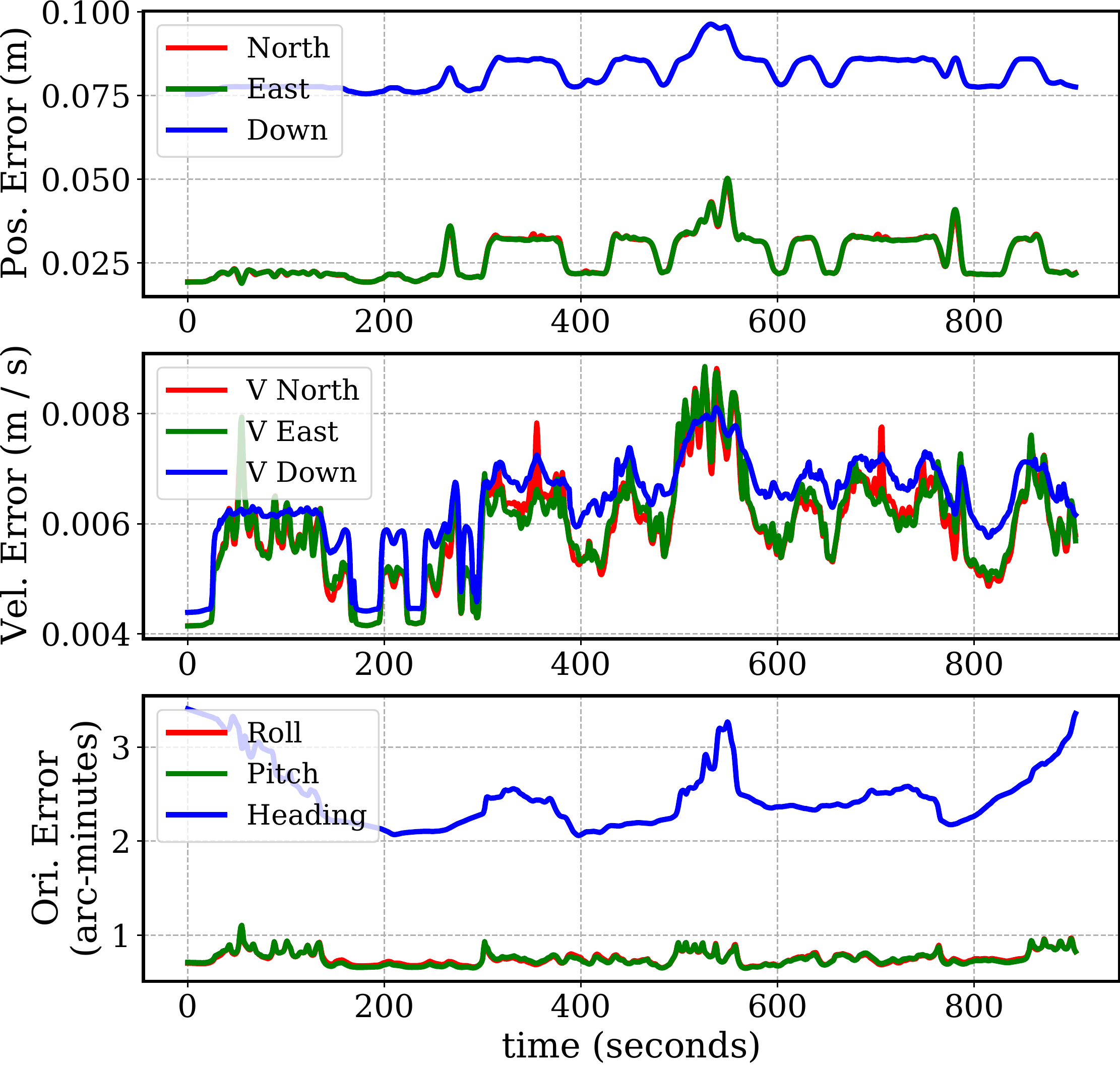}
	\caption{Post-processed RMS position, velocity, and orientation residual error vs. time reported by Applanix's POSPac software for a sequence collected on 2021-09-07.}
	\label{fig:gterr}
\end{figure}

\begin{equation}
	\begin{bmatrix} u \\ v \end{bmatrix} = \mathbf{D}~\mathbf{P}~\frac{1}{z}~\mathbf{x}_c
\end{equation}

\begin{equation}
	\text{where} ~ \mathbf{D} = \begin{bmatrix} 1 & 0 & 0 & 0 \\ 0 & 1 & 0 & 0 \end{bmatrix},  \mathbf{P} = \begin{bmatrix} f_u & 0 & c_u & 0 \\ 0 & f_v & c_v & 0 \\ 0 & 0 & 1 & 0 \\ 0 & 0 & 0 & 1 \end{bmatrix}.
\end{equation}

\subsection{Sensor Extrinsics}

\begin{figure}[t]
	\centering
	\subfigure[Perspective View]{\includegraphics[width=1.0\columnwidth]{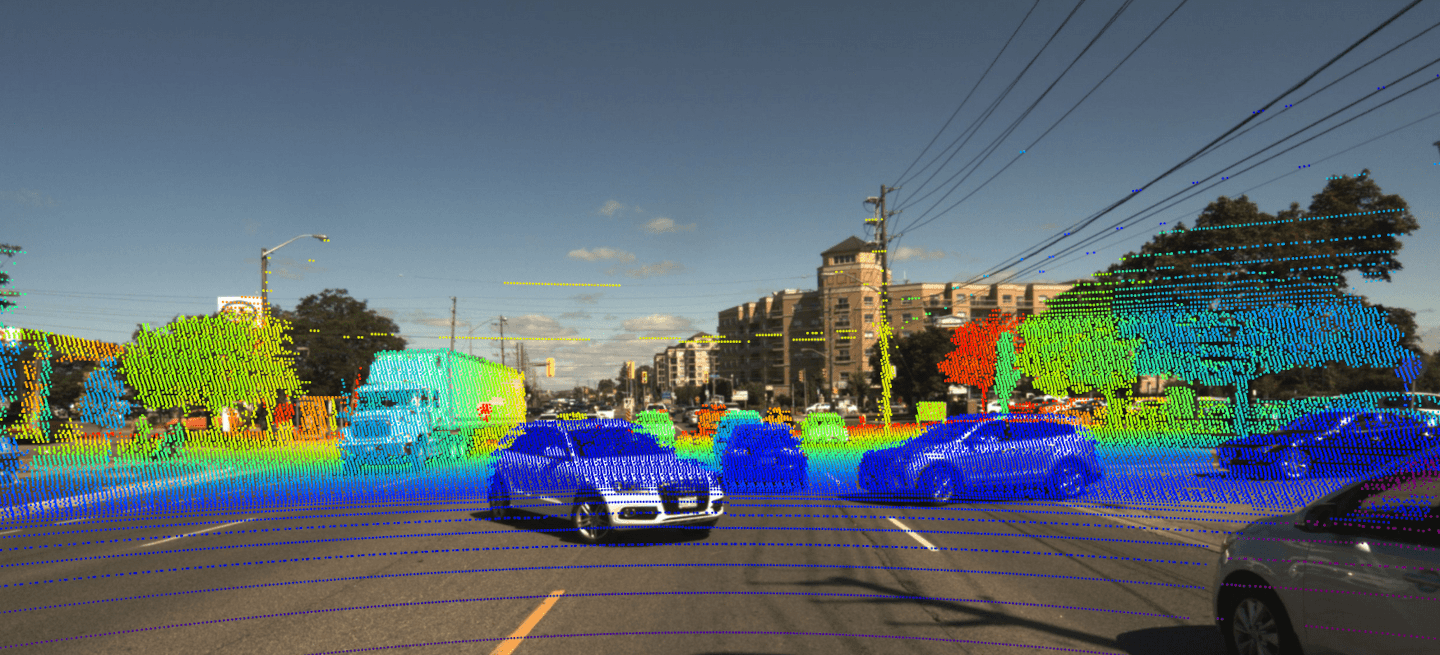}}
	\subfigure[Colored Pointcloud]{\includegraphics[width=1.0\columnwidth]{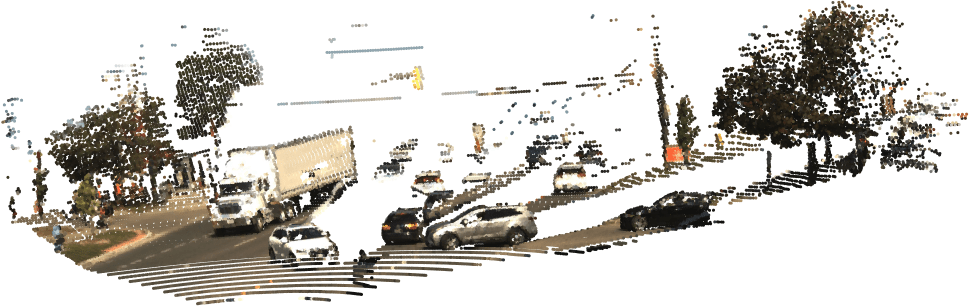}}
	\caption{Lidar points projected onto a camera image using the camera-lidar calibration. (a) Lidar points are colored based on their longitudinal distance from the vehicle. (b) Lidar points are given RGB color values based on their projected location on the camera image.}
	\label{fig:camvel}
\end{figure}

\begin{figure}[t]
	\centering
	\includegraphics[width=0.95\columnwidth]{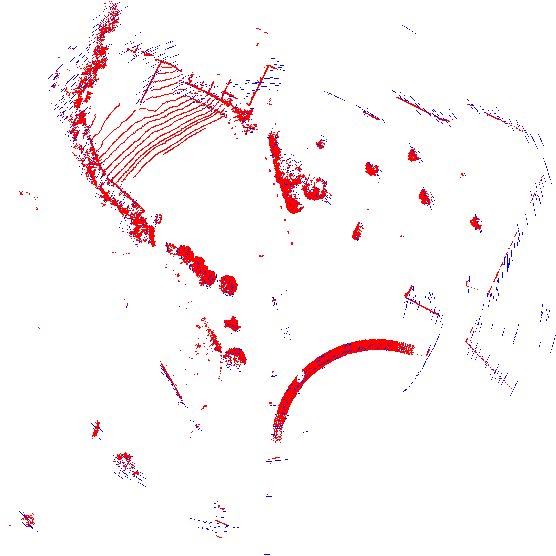}
	\caption{Lidar measurements are drawn in red using a bird's eye view projection with the ground plane removed. Radar targets are first extracted from the raw radar data and then are drawn as blue pixels. The two sensors have been aligned using the radar-to-lidar calibration.}
	\label{fig:radvel}
\end{figure}

The extrinsic calibration between the camera and lidar is obtained using MATLAB's camera to lidar calibrator \citep{matlab}. The results of this calibration are illustrated in Figure~\ref{fig:camvel}. To calibrate the rotation between the lidar and radar, we use correlative scan matching via the Fourier Mellin transform \citep{checchin_fsr10}. Several lidar-radar pairs were collected while the vehicle was stationary at different locations. The final rotation estimate is obtained by averaging the results from several measurement pairs \citep{radar-lidar-calib}. The translation between the lidar and radar is obtained from the computer assisted design (CAD) model of the roof rack. The results of the radar-to-lidar calibration are shown in Figure~\ref{fig:radvel}. The extrinsics between the lidar and the Applanix reference frame were obtained using Applanix's in-house calibration tools. Their tool outputs this relative transform as a by-product of a batch optimization aiming to estimate the most likely vehicle path given a sequence of lidar pointclouds and post-processed GNSS/IMU measurements. All extrinsic calibrations are provided as 4x4 homogeneous transformation matrices under the {\ttfamily{calib/}} folder.



\section{3D Annotations}

\begin{figure}[b]
	\centering
	\includegraphics[width=1.0\columnwidth]{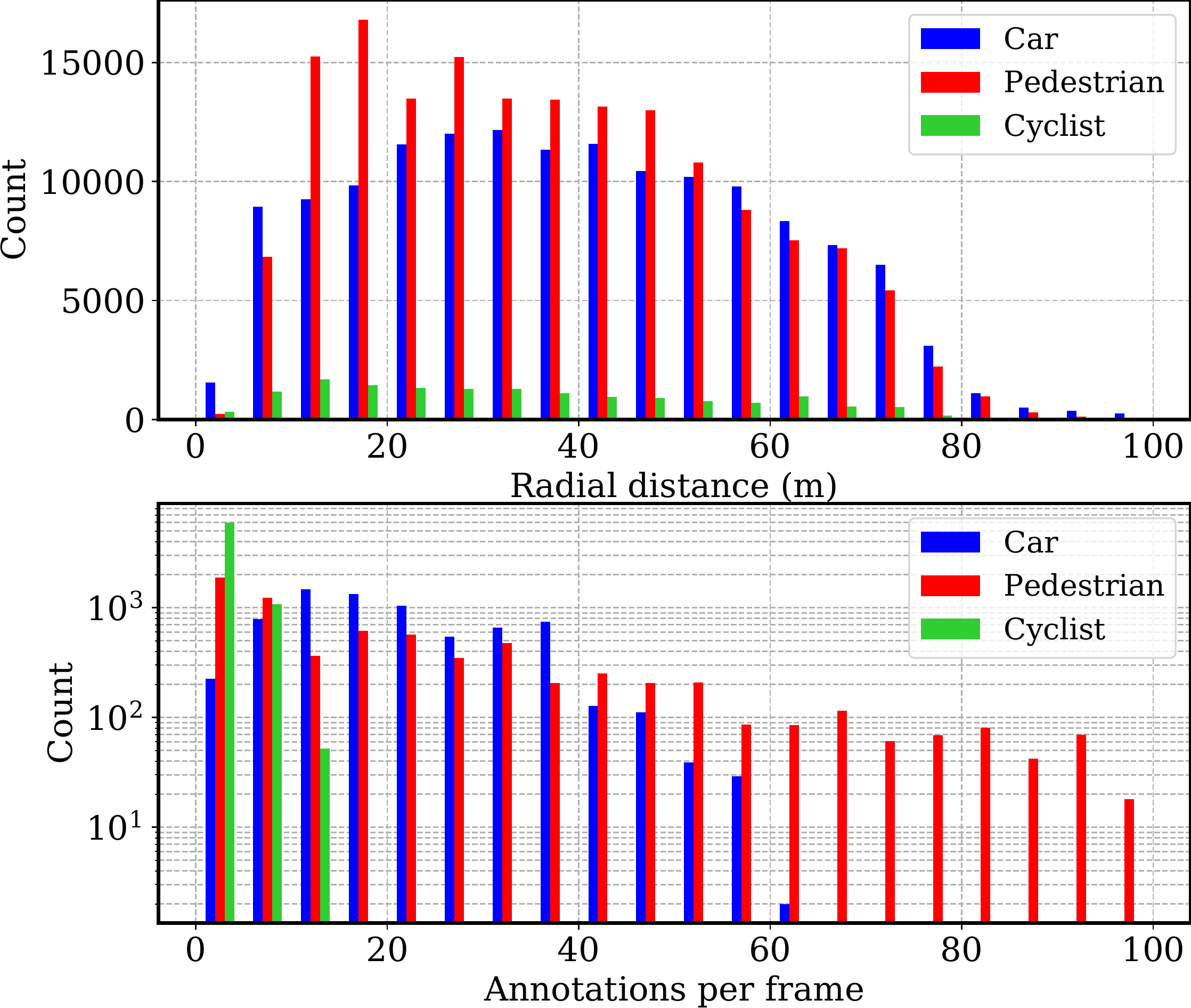}
	\caption{3D annotation statistics for Boreas-Objects-V1.}
	\label{fig:objstats}
\end{figure}

We provide a set of 3D bounding box annotations for a subset of the Boreas dataset, obtained in sunny weather. We refer to this as the Boreas-Objects-V1 dataset. Annotations were obtained using the Scale.ai data annotation service \citep{scale}. In total, 7111 lidar frames were annotated at 5Hz, resulting in 326,180 unique 3D box annotations. Since the lidar data was collected at 10Hz, the annotations may be interpolated between frames to double the number of annotated frames at a slightly lower fidelity. The data is divided into 53 continuous scenes where each scene is 20-70 seconds in duration. The scenes are then divided into 37 training scenes and 16 test scenes where the ground truth labels have been withheld for the benchmark. Figure~\ref{fig:objstats} displays two statistics for our annotations.

\begin{figure*}[ht]
	\centering
	\includegraphics[width=1.0\textwidth]{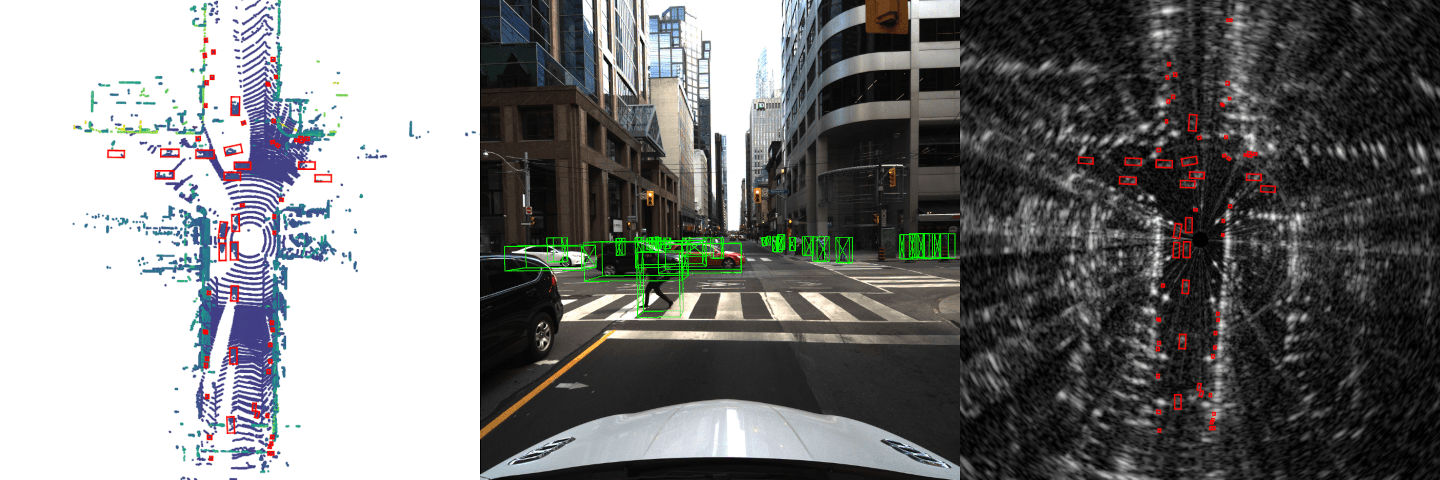}
	\caption{Examples of 3D annotations in the Boreas-Objects-V1 dataset.}
	\label{fig:objexamp}
\end{figure*}

We use the same folder structure as in Figure~\ref{fig:data} but with an additional folder, {\ttfamily{labels/}}. Similar to KITTI, annotations for a particular frame are stored in a text file with the same filename (timestamp) as the lidar frame. Each row of a label file corresponds to a different 3D box annotation with the format: $[\text{uuid}, \text{type}, d_x, d_y, d_z, x, y, z, \text{yaw}]$. The uuid is a unique ID for a particular object track that is consistent across frames within a particular scene. The type is the semantic class for an object that can be one of: \{Car, Cyclist, Pedestrian, Misc\}. The Car class includes coupes, sedans, SUVs, vans, pick-up trucks, and ambulances. The Cyclist class includes people riding motorcycles, but excludes parked bicycles. The Misc class includes other vehicle types such as buses, industrial trucks, streetcars, and trains. Objects are labelled within a rectangular area centered on the lidar +/- 75m in both dimensions. Bounding box locations $(x, y, z)$ and orientations (yaw) are given with respect to the lidar frame. $(d_x, d_y, d_z)$ represent the bounding box dimensions (length, width, and height). Figure~\ref{fig:objexamp} shows an example of what our 3D object annotations look like for lidar, camera, and radar.

\section{Benchmark Metrics}

At launch, we plan to support online leaderboards for odometry, metric localization, and 3D object detection. For odometry, we use the same metrics as the KITTI dataset \citep{kitti}. The KITTI odometry metrics average the relative position and orientation errors over every sub-sequence of length (100m, 200m, 300m, ..., 800m). This results in two metrics, a translational drift reported as a percentage of path length and a rotational drift reported as degrees per metre travelled. For 3D object detection, we also defer to the KITTI dataset by reporting the mean average precision (mAP) on a per-class basis. For cars, a 70\% overlap counts as a true positive, and for pedestrians, 50\%. These ratios are used as they are the same as what was used in the KITTI dataset. We do not divide our dataset based on difficulty levels.


The purpose of our metric localization leaderboard is to benchmark mapping and localization pipelines. In this scenario, we envision a situation where one or more repeated traversals of the Glen Shields route are used to construct a map offline. Any and all data  from the training sequences may be used to construct a map in any fashion.

Then, during a test sequence, the goal is to perform metric localization between the live sensor data and the pre-built map. Localization approaches may make use of temporal filtering and can leverage the IMU if desired but GNSS information will not be available. The goal of this benchmark is to simulate localizing a vehicle in real-time and as such methods may not use future sensor information in an acausal manner.

Our goal is to support both global and relative map structures. Only one of the training sequences will be specified as the map sequence used by the benchmark. For 3D localization, users must choose either the lidar or camera as the reference sensor. For 2D localization, only the radar frames are used as a reference. For each (camera|lidar|radar) frame $s_2$ in the test sequence, users will specify the ID (timestamp) of the (camera|lidar|radar) frame $s_1$ in the map sequence that they are providing a relative pose with respect to: $\hat{\mathbf{T}}_{s_1,s_2}$. We then compute root-mean squared error (RMSE) values for the translation and rotation as follows:

\begin{align}
	\mathbf{T}_e &=  \mbf{T}_{a,s_1} \mathbf{T}_{s_1,s_2} \hat{\mathbf{T}}_{s_1,s_2}^{-1} \mbf{T}_{a,s_1}^{-1} = \begin{bmatrix} \mathbf{C}_e & \mathbf{r}_e \\ \mathbf{0}^T & 1  \end{bmatrix} \\
	\mathbf{r}_e &= \begin{bmatrix} x_e & y_e & z_e \end{bmatrix}^T \\
	\phi_e &= \arccos \left( \frac{\text{tr}~\mathbf{C}_e - 1}{2}  \right)
\end{align}

where $\mathbf{T}_{s_1,s_2}$ is the known ground truth pose, and $\mbf{T}_{a,s_1}$ is the calibrated transform from the sensor frame to the applanix frame (x-right, y-forwards, z-up). $x_e, y_e, z_e$ are then the lateral, longitudinal, and vertical errors respectively. We calculate RMSE values for $x_e, y_e, z_e, \phi_e$.

Users will also have the option of providing $6 \times 6$ covariance matrices $\mbs{\Sigma}_i$ for each localization estimate. A pose with uncertainty is described as $\mbf{T} = \exp(\x^\wedge) \overline{\mbf{T}}$ where $\x \sim \mathcal{N} \left(\mbf{0}, \mbs{\Sigma} \right)$ \citep{barfoot2017state}. Given $\hat{\mathbf{T}}_{i} = \hat{\mathbf{T}}_{s_1,s_2}(t_i)$, we compute an average consistency score $c$ for the localization and covariance estimates:


\begin{align}
	\x_i &= \ln \left( \mathbf{T}_{i} \hat{\mathbf{T}}_{i}^{-1} \right)^\vee =  \bbm \rho_1 & \rho_2 & \rho_3 & \psi_1 & \psi_2 & \psi_3 \ebm^T \\
	c &=  \left( \sum_{i=1}^N \frac{\x_i^T \mbs{\Sigma}_i^{-1} \x_i}{N \text{dim}(\x_i)} \right)^{1/2}
\end{align}

A consistency score close to 1 is ideal. $c < 1$ means that the method is over-confident, $c > 1$ means that the method is conservative. Note that the above metrics will be averaged across the test sequences.

\section{Development Kit}
As part of this dataset, we provide a development kit for new users to get started. The primary purpose of the devkit is to act as a wrapper around the dataset to be used in Python. This allows users to query frames and the associated ground truth for either odometry, localization, or 3D object detection. We also provide convenience methods for removing motion distortion from pointclouds, working with polar radar scans, and converting to and from Lie algebra and Lie group representations. The devkit also provides several ways to visualize sensor data. We also provide introductory tutorials in Jupyter notebooks that include projecting lidar onto a camera frame and visualizing 3D boxes. Evaluation scripts used by our benchmark will be stored in the devkit, allowing users to validate their algorithms before submission to the benchmark. The development kit can be found at \href{https://www.boreas.utias.utoronto.ca}{boreas.utias.utoronto.ca}.

\section{Conclusion}

In this paper, we presented Boreas, a multi-season autonomous driving dataset that includes over 350km of driving data collected over the course of one year. The dataset provides a unique high-quality sensor suite including a Velodyne Alpha-prime (128-beam) lidar, a 5MP camera, a 360$^\circ$ Navtech radar, and accurate ground truth poses obtained from an Applanix POS~LV with an RTX subscription. We also provide 3D object labels for a subset of the Boreas data obtained in sunny weather. The primary purpose of this dataset is to enable further research into long-term localization across seasons and adverse weather conditions. Our website will provide an online leaderboard for odometry, metric localization, and 3D object detection.

\begin{acks}
	We would like to thank Goran Basic for his help in designing and assembling the roof rack for Boreas. We also thank General Motors for their donation of the Buick vehicle. The Amazon Open Data Sponsorship program supports this project by hosting the Boreas dataset. This work was also partially financially supported by a Natural Sciences and Engineering Research Council (NSERC) grant.
\end{acks}

\bibliographystyle{SageH}
\bibliography{bib/refs}

\end{document}